\documentclass[journal]{IEEEtran}

\usepackage{amsmath}
\usepackage{graphicx}
\usepackage{amssymb}
\usepackage{tikz}
\usetikzlibrary{arrows.meta,positioning,fit,calc}
\usepackage{algorithm}
\usepackage{algorithmic}\usepackage{placeins}
\usepackage{cite}

\begin{document}

\title{A Schema Bounded Language Model for Refining Robot Policies Without Destabilizing Local Learning}

\author{
 \IEEEauthorblockN{1\textsuperscript{st} Chongwen Dong}\\
 \IEEEauthorblockA{
   \textit{Department of Mechanical Engineering}\\
   \textit{Northern Arizona University}\\
     15600 S. McConnell Dr.,
           Flagstaff, AZ 86011\\}
 \and
 \IEEEauthorblockN{2\textsuperscript{nd} Mithun Paul Saint-Germain and Pinjari Asif}\\
 \IEEEauthorblockA{
   \textit{School of Informatics, Computing, and Cyber Systems}\\
   \textit{Northern Arizona University}\\
     1295 Knoles Dr.,
     Flagstaff, AZ 86011\\}
 \and
 \IEEEauthorblockN{3\textsuperscript{rd} Carlo daCunha}\\
 \IEEEauthorblockA{\textit{Helen and John C. Hartmann}\\
 \textit{Department of Electrical Engineering} and \\
 \textit{School of Applied Engineering and Technology}\\
 \textit{New Jersey Institute of Technology}\\
	323 Martin Luther King Jr. Blvd.,
	Newark, NJ 07102\\
	carlo.dacunha@njit.edu}
}

\markboth{IEEE Journal Manuscript}%
{Carlo and Mithun: Two-Level Cross-LLM Robot Policy Refinement}

\maketitle

\begin{abstract}
This paper addresses navigation by composite heterogeneous robots in a decentralized system when policy reasoning and local control operate at different update levels. In a NetLogo--Python implementation, three robots share motion dynamics but use different LLM backends. Each robot independently combines a large language model (LLM) policy agent, an Upper Confidence Bound (UCB) bandit, and a Double Deep Q-Network (Double DQN) controller; no central LLM generates team actions. LLM inference is confined to round-level policy generation and refinement rather than tick-level action selection. The robots perform cross-LLM communication through a shared round summary containing policies, outcomes, and learning feedback. UCB performs refinement-mode selection, and the policy-conditioned Double DQN performs tick-level action selection from navigation variables, active policy parameters, and the LLM action prior. Each of the four configurations was evaluated over 30 rounds. In the fixed simulation, the complete configuration reached the goal in all 90 correlated robot--round records and achieved the lowest median completion time (42 ticks) and P90 (73.2 ticks); its median was 25.0--39.1\% lower than those of the other configurations. These observations provide descriptive, configuration-level evidence from the evaluated configurations.
\end{abstract}


\IEEEpeerreviewmaketitle

\section{Introduction}
\IEEEPARstart{H}{eterogeneous} multi-robot systems combine agents with different physical or decision capabilities to perform shared tasks. Large language models (LLMs) extend these systems with language-mediated task decomposition, communication, and policy adaptation \cite{roco,scalablellm}. For continual navigation, however, policy reasoning and motion control operate at different update levels. An LLM can generate or refine a policy at a task boundary, whereas a local controller must select an action at every environment tick. Calling an LLM for each action would place policy reasoning inside the local control loop. The resulting design problem is to retain LLM reasoning at the policy level while providing a separate mechanism for tick-level execution.

Existing LLM-based multi-robot frameworks have investigated dialogue-based coordination, capability-aware task allocation, decentralized planning, and feedback-driven policy revision \cite{modular,smartllm,mhrc,coherent,sasprompt}. Hierarchical robotic systems have also separated high-level language reasoning from lower-level execution \cite{genswarm,ellmer}. This current work is motivated by these aforementioned works, but instead asks a more specific system-design question: how can persistent robot-specific LLM policies be refined from shared completed-round information while local robot learning supplies every motion action? This question requires an explicit interface among policy ownership, cross-LLM communication, round-level policy refinement, and tick-level action selection.

This paper introduces a solution for this problem: a decentralized two-tier architecture one with robot-local policy ownership, and the second one guided by LLMs. Specifically, in the first tier, a local policy guidance is combined with a Deep Q-Network (DQN). This tier recieves input from second tier which has an LLM shared board which is in-turn guided by Upper Confidence Bound (UCB) algorithm. Specifically the contributions of this work are as follows:

\begin{enumerate}
    \item A one-LLM-per-robot architecture for composite heterogeneous robots that keeps policy ownership, UCB-guided refinement, and Double DQN control local to each robot;
    \item A temporally decoupled LLM policy-generation and refinement mechanism that confines LLM inference to round boundaries, while robot-local controllers perform tick-level action selection without per-action LLM invocation;
    \item A shared-board-mediated cross-LLM communication mechanism that exchanges completed-round policies, outcomes, rationales, DQN feedback, and UCB advice while preserving robot-local policy ownership;
\end{enumerate}

\section{Related Work}
LLM-based multi-agent and multi-robot systems organize decision authority in centralized, decentralized, or hybrid forms. COHERENT uses a centralized task assigner to decompose long-horizon tasks for heterogeneous robot executors \cite{coherent}. In contrast, MHRC places LLM reasoning at individual heterogeneous robots and coordinates them through local observations and inter-robot messages \cite{mhrc}. Chen et al. compare centralized, decentralized, and hybrid communication frameworks, showing that topology changes both coordination performance and token use \cite{scalablellm}. RoCo adopts a hybrid organization: robot-specific LLM agents deliberate over subtask and waypoint plans, while a centralized multi-arm motion planner produces executable trajectories \cite{roco}. SMART-LLM addresses capability-aware task allocation \cite{smartllm}. Modular embodied-agent frameworks use LLMs to construct cooperative agent systems \cite{modular}, whereas EMOS incorporates embodiment descriptions into heterogeneous multi-robot planning \cite{emos}. Collectively, these studies show that shared communication does not by itself determine whether a system is decentralized; the location of decision authority and persistent state is also decisive. In the architecture studied here, the shared board transports completed-round information, while each robot retains its own policy and learning states.

A second distinction is the temporal level at which language reasoning enters control. SAS-Prompt treats an LLM as a numerical policy optimizer that revises robot behavior from trajectory traces \cite{sasprompt,CarloAIRC}. GenSwarm generates deployable code policies for multi-robot tasks \cite{genswarm}. ELLMER separates high-level LLM reasoning and generated code from sensorimotor feedback during execution \cite{ellmer}. YOLO-MARL makes the separation between language reasoning and local execution more explicit: it queries an LLM before multi-agent reinforcement learning and subsequently executes decentralized neural policies without repeated LLM inference \cite{yolomarl}. These methods range from language-generated task or motion specifications to persistent policies executed by non-LLM controllers. Relative to these approaches, the proposed system uses dedicated LLMs to generate and refine robot-specific policies at round boundaries, while robot-local controllers perform tick-level action selection.

Reinforcement learning provides several interfaces between language-level guidance and executable multi-agent behavior. Language-conditioned offline RL embeds natural-language commands into decentralized multi-robot navigation policies \cite{languageoffline}. YOLO-MARL uses LLM-generated planning functions to guide subsequent MARL policy learning \cite{yolomarl}. LAMARL instead uses LLM-generated prior policies and reward functions to guide cooperative multi-robot training \cite{lamarl}. Yoshida and Sueoka combine LLM-based high-level policy selection with MARL-based swarm control \cite{yoshida}, while hierarchical analyses model an LLM planner above a lower-level RL actor \cite{hewords}. In contrast to approaches where RL learns the principal policy from language-provided specifications, RL is auxiliary to the LLM policy process in this work. The LLM generates and refines the round-level policy. DQN and Double Q-learning provide the algorithmic foundations for the robot-local Double DQN, which selects tick-level actions and supplies execution feedback \cite{mnih,doubleq}. UCB1 contributes the exploration--exploitation rule used only to select the refinement mode for the next LLM update \cite{auer}. The resulting interface combines persistent one-LLM-per-robot policy ownership, completed-round cross-LLM communication, and robot-local learning assistance without assigning team-level policy generation or robot action selection to a central LLM.

\section{Framework}
\label{sec:framework}

\subsection{System Model and Overall Framework}
This paper introduces a solution for this problem: a decentralized two-tier architecture one with robot-local policy ownership, and the second one guided by LLMs. Specifically, in the first tier, a local policy guidance is combined with a Deep Q-Network (DQN). This tier recieves input from second tier which has an LLM shared board which is in-turn guided by Upper Confidence Bound (UCB) algorithm. In detail, a dedicated LLM generates and refines the robot-specific policy at round boundaries, while the local controller selects tick-level actions based on inputs from the LLMs. That is, policy ownership and learning states remain local to each robot; no central LLM generates a team policy or controls robot actions. Further, round-level communication and policy updates follow the event-driven schedule (detailed in Section~\ref{sec:round_update}). Further, we introduce a shared board that aggregates completed-round policies, outcomes, rationales, DQN feedback, and UCB advice. Each LLM reads the same board but retains its robot's policy history and generates only that robot's next policy. UCB selects a refinement mode, and the policy-conditioned Double DQN selects tick-level actions and returns execution feedback. In this paper, \emph{heterogeneous robots} denotes model- and decision-layer heterogeneity within this composite architecture; robot embodiment and motion dynamics remain fixed.

Consider \(N\) mobile robots \(\mathcal{R}=\{r_1,\ldots,r_N\}\) that repeat the same goal-reaching task for \(K\) rounds, with at most \(T_{\max}\) control ticks per round. The notation indexes robot-local quantities by \(N\); the implementation and experiments use \(N=3\), and no claim is made about team-size scalability. At tick \(t\) of round \(k\), NetLogo maintains each turtle's position, heading, terminal flags, and finish tick. These variables determine the current goal distance and signed heading error. Success requires the turtle's current patch to equal the goal patch.

A control tick is one pass through the local action phase for every active robot. NetLogo increments the global counter after this phase. An arrival or timeout detected after an executed action is therefore recorded as \texttt{ticks + 1}, whereas a turtle already on the goal patch retains the current counter because it executes no new action. Let \(y_i^k\in\{0,1\}\) denote success and \(T_i^k\) the recorded completion tick. A timeout gives \(y_i^k=0\) and \(T_i^k=T_{\max}\); \(d_i^0\) denotes the fixed start-to-goal distance used by the round-level credit.

Each robot is implemented as a four-component composite agent,
\begin{equation}
r_i=
\left\langle
\mathcal{T}_i,\mathcal{M}_i,\mathcal{U}_i,\mathcal{D}_i
\right\rangle ,
\end{equation}
where \(\mathcal{T}_i\) is the NetLogo turtle, \(\mathcal{M}_i\) is its dedicated LLM policy agent, \(\mathcal{U}_i\) is its UCB1 bandit, and \(\mathcal{D}_i\) is its Double DQN controller. The three robots are assigned Llama-3.3-70B-Instruct-quantized, Phi-4, and Meta-Llama-3.1-70B-Instruct-quantized, respectively. Each robot retains its own LLM policy history, UCB values and counts, online and target networks, replay buffer, exploration state, and action statistics.

The evaluated heterogeneity is at the model and decision layers. The turtles share embodiment, motion rules, and action categories, but use different LLM backends and develop robot-specific policies, UCB histories, and DQN trajectories. Fixed initial poses are experimental conditions rather than morphological heterogeneity. The system is decentralized with respect to decision authority and persistent control state: no central LLM creates a team policy or selects robot actions. The shared text board, NetLogo world, and Python runtime are simulation infrastructure; they exchange or execute robot-local information without replacing the individual controllers.

The code executes in two coupled loops. During setup, the Python layer initializes the NetLogo task and the robot-local UCB and Double DQN states, then generates the three initial LLM policies concurrently. During a rollout, each active turtle provides its local state, its Double DQN selects one action category, NetLogo applies the active policy parameters, and the resulting reward and next state are stored for robot-local training. When all turtles are terminal, the Python layer records the round, credits any UCB arm associated with the evaluated policy, selects the next refinement mode, and updates best-so-far records. At an eligible round boundary, it constructs the shared board, queries the dedicated LLMs concurrently, sanitizes and loads the returned robot-specific policies, and resets the task for the next rollout. Section~\ref{sec:round_update} defines the exact LLM-call schedule and failure handling.

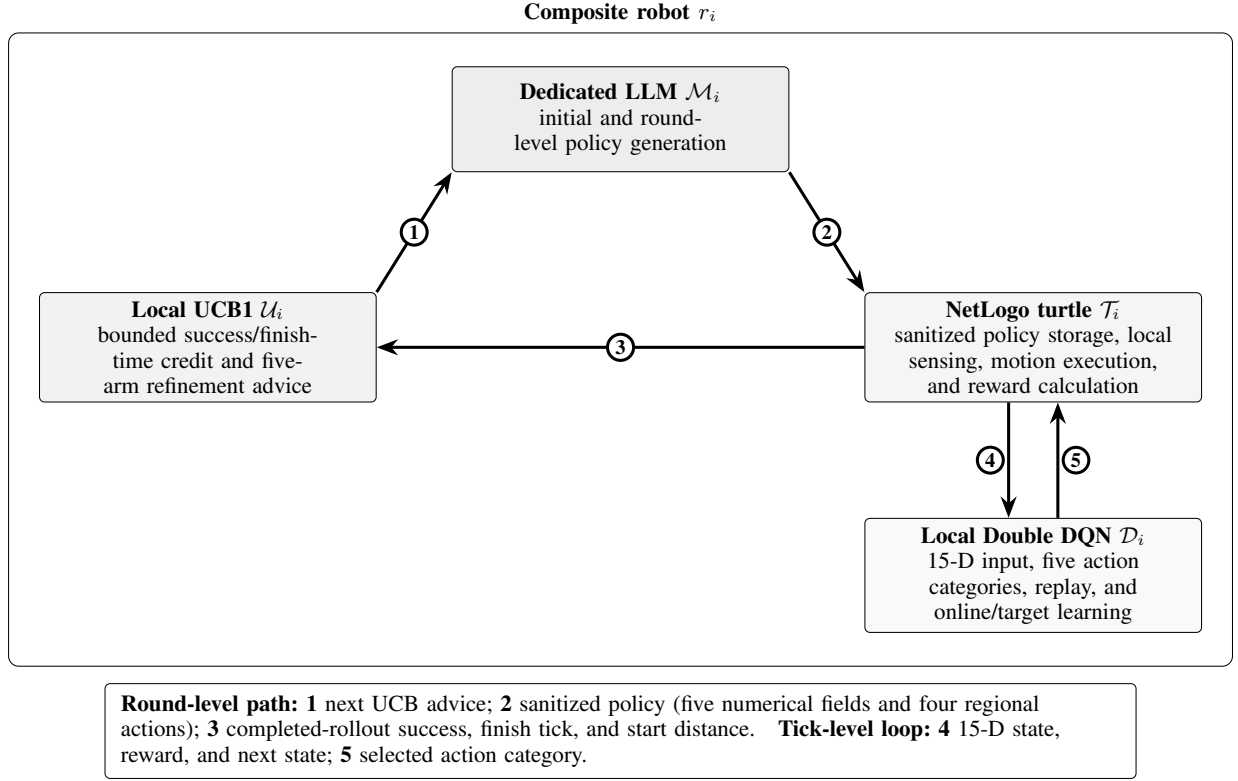
\begin{figure*}[t]
\centering
\resizebox{0.90\textwidth}{!}{%
\begin{tikzpicture}[
font=\small,
module/.style={
draw,
rounded corners=2pt,
minimum width=4.65cm,
minimum height=1.45cm,
text width=4.25cm,
align=center,
fill=gray!8
},
robotbox/.style={
draw,
rounded corners=4pt,
inner xsep=0.42cm,
inner ysep=0.46cm
},
flow/.style={-Stealth,very thick},
tag/.style={circle,draw,fill=white,inner sep=1.2pt,font=\bfseries\footnotesize},
legend/.style={draw,rounded corners=2pt,fill=white,align=left,
text width=13.8cm,inner xsep=0.22cm,inner ysep=0.16cm}
]
\node[module,fill=gray!14] (llm) at (0,3.15)
{\textbf{Dedicated LLM $\mathcal{M}_i$}\\initial and round-level
policy generation};
\node[module,fill=gray!10] (ucb) at (-5.7,0)
{\textbf{Local UCB1 $\mathcal{U}_i$}\\bounded success/finish-time credit and
five-arm refinement advice};
\node[module,fill=gray!5] (dqn) at (5.7,-3.15)
{\textbf{Local Double DQN $\mathcal{D}_i$}\\15-D input, five action categories,
replay, and online/target learning};
\node[module,fill=gray!8] (turtle) at (5.7,0)
{\textbf{NetLogo turtle $\mathcal{T}_i$}\\sanitized policy storage, local
sensing, motion execution, and reward calculation};

\node[robotbox, fit=(llm)(ucb)(dqn)(turtle),
label={[font=\bfseries\small]above:Composite robot $r_i$}] {};

\draw[flow] (ucb.north east) -- node[tag,pos=0.50] {1} (llm.south west);
\draw[flow] (llm.south east) -- node[tag,pos=0.50] {2} (turtle.north west);
\draw[flow] (turtle.west) -- node[tag,pos=0.50] {3} (ucb.east);
\draw[flow] ([xshift=-0.34cm]turtle.south) --
node[tag,pos=0.50,left=1.5pt] {4} ([xshift=-0.34cm]dqn.north);
\draw[flow] ([xshift=0.34cm]dqn.north) --
node[tag,pos=0.50,right=1.5pt] {5} ([xshift=0.34cm]turtle.south);

\node[legend] (legend) at (0,-5.30) {%
\textbf{Round-level path:} \textbf{1} next UCB advice;
\textbf{2} sanitized policy (five numerical fields and four regional actions);
\textbf{3} completed-rollout success, finish tick, and start distance.\quad
\textbf{Tick-level loop:} \textbf{4} 15-D state, reward, and next state;
\textbf{5} selected action category.};
\end{tikzpicture}
}
\caption{Information flow within composite robot $r_i$. Numbered arrows keep the
round-level policy path visually separate from the tick-level Double DQN loop;
all four components and their persistent states are owned by the same robot.}
\label{fig:composite-architecture}
\end{figure*}

Fig.~\ref{fig:composite-architecture} separates the round-level policy path from the robot-local tick-level loop.

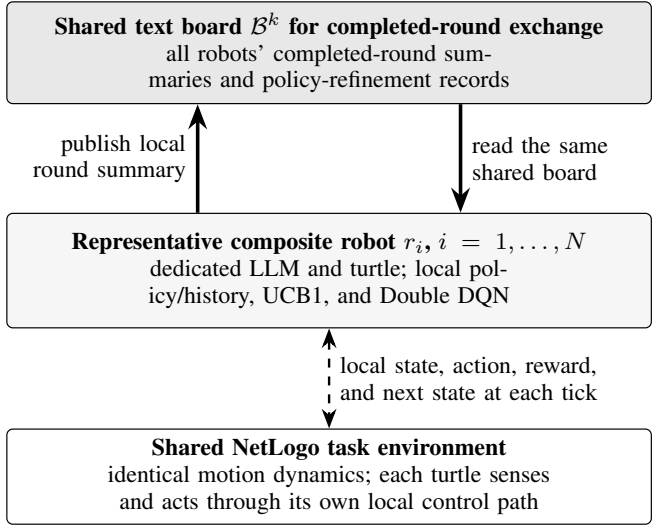
\begin{figure}[t]
\centering
\resizebox{0.98\columnwidth}{!}{%
\begin{tikzpicture}[
font=\footnotesize,
board/.style={
draw,
rounded corners=2pt,
minimum width=7.6cm,
minimum height=1.20cm,
text width=7.20cm,
align=center,
fill=gray!18
},
robot/.style={
draw,
rounded corners=2pt,
minimum width=7.6cm,
minimum height=1.35cm,
text width=7.20cm,
align=center,
fill=gray!7
},
environment/.style={
draw,
rounded corners=2pt,
minimum width=7.6cm,
minimum height=1.05cm,
text width=7.20cm,
align=center
},
roundflow/.style={-Stealth,very thick},
tickflow/.style={Stealth-Stealth,thick,dashed}
]
\node[board] (board)
{\textbf{Shared text board $\mathcal{B}^{k}$ for completed-round exchange}\\
all robots' completed-round summaries and policy-refinement records};

\node[robot, below=1.28cm of board] (robot)
{\textbf{Representative composite robot $r_i$, $i=1,\ldots,N$}\\
dedicated LLM and turtle; local policy/history, UCB1, and Double DQN};

\node[environment, below=1.18cm of robot] (env)
{\textbf{Shared NetLogo task environment}\\identical motion dynamics; each
turtle senses and acts through its own local control path};

\draw[roundflow] ([xshift=-1.55cm]robot.north) --
node[left,align=right] {publish local\\round summary}
([xshift=-1.55cm]board.south);
\draw[roundflow] ([xshift=1.55cm]board.south) --
node[right,align=left] {read the same\\shared board}
([xshift=1.55cm]robot.north);

\draw[tickflow] (robot.south) --
node[right,align=left] {local state, action, reward,\\and next state at each tick}
(env.north);
\end{tikzpicture}
}
\caption{Representative communication topology. The board is common to all
robots, whereas policy generation, UCB1, Double DQN, and turtle execution remain
robot-local. Solid arrows denote round-level information exchange; the dashed arrow denotes
the robot-local tick-level environment loop.}
\label{fig:decentralized-communication}
\end{figure}

Cross-robot information is limited to completed-round records published through \(\mathcal{B}^{k}\). Fig.~\ref{fig:decentralized-communication} shows this boundary for a representative robot; the same local topology is instantiated for all three robots.

\subsection{Round-Level Policy Refinement and Cross-LLM Communication}
\label{sec:round_update}

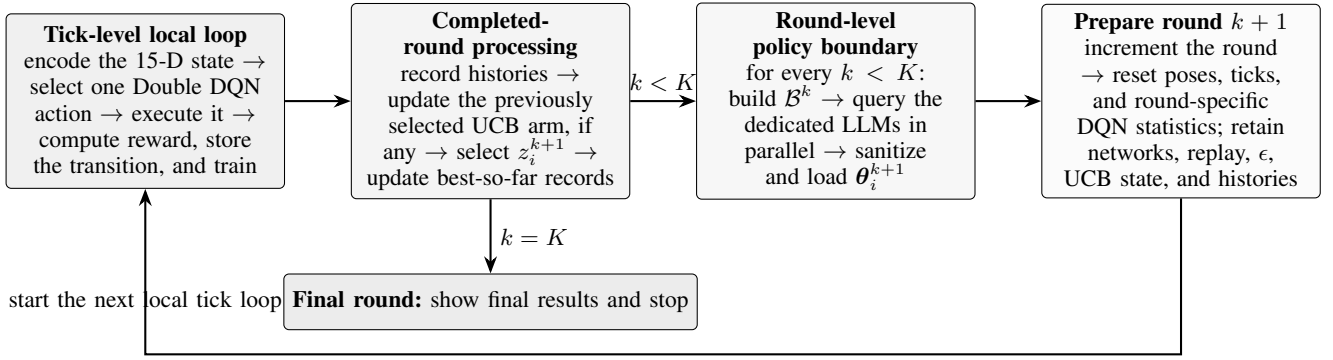
\begin{figure*}[t]
\centering
\resizebox{0.97\textwidth}{!}{%
\begin{tikzpicture}[
font=\small,
stagebox/.style={
draw,
rounded corners=2pt,
minimum width=3.75cm,
minimum height=2.35cm,
text width=3.42cm,
align=center,
fill=gray!6
},
stopbox/.style={
draw,
rounded corners=2pt,
minimum width=3.35cm,
minimum height=0.75cm,
align=center,
fill=gray!15
},
flow/.style={-Stealth,thick}
]
\node[stagebox, fill=gray!12] (tick) at (0,0)
{\textbf{Tick-level local loop}\\
encode the 15-D state $\rightarrow$ select one Double DQN action
$\rightarrow$ execute it $\rightarrow$ compute reward, store the transition,
and train};

\node[stagebox, fill=gray!9] (round) at (4.65,0)
{\textbf{Completed-round processing}\\
record histories $\rightarrow$ update the previously selected UCB arm, if any
$\rightarrow$ select $z_i^{k+1}$ $\rightarrow$ update best-so-far records};

\node[stagebox, fill=gray!6] (policy) at (9.30,0)
{\textbf{Round-level policy boundary}\\
for every $k<K$: build $\mathcal{B}^{k}$ $\rightarrow$ query the
dedicated LLMs in parallel $\rightarrow$ sanitize and load
$\boldsymbol{\theta}_{i}^{k+1}$};

\node[stagebox, fill=gray!3] (reset) at (13.95,0)
{\textbf{Prepare round $k+1$}\\
increment the round $\rightarrow$ reset poses, ticks, and round-specific DQN
statistics; retain networks, replay, $\epsilon$, UCB state, and histories};

\node[stopbox, below=1.00cm of round] (stop)
{\textbf{Final round:} show final results and stop};

\draw[flow] (tick) -- (round);
\draw[flow] (round) -- node[above] {$k<K$} (policy);
\draw[flow] (policy) -- (reset);
\draw[flow] (round.south) -- node[right] {$k=K$} (stop.north);
\draw[flow] (reset.south) -- ++(0,-2.10) -|
node[pos=0.72,below] {start the next local tick loop} (tick.south);
\end{tikzpicture}
}
\caption{Implemented control and update cycle. Once all turtles are terminal,
tick-level Double DQN learning is followed by completed-round history and UCB updates. Shared-board
construction and parallel LLM regeneration occur after every non-final round.
The final round terminates before board construction or another LLM call.}
\label{fig:update-cycle}
\end{figure*}

The implementation uses an event-driven two-level schedule. Each dedicated LLM generates one initial policy before round 1. After every completed nonfinal round \(k<K\), UCB evaluates the rollout and selects a refinement mode, after which that robot's LLM is called once to generate the policy for round \(k+1\). Evaluating \(K\) rounds therefore requires exactly \(K\) LLM calls per robot: one initialization call and \(K-1\) refinement calls. No LLM is called within the tick-level control loop or after the final round. The two levels denote round-level policy adaptation and tick-level local control, rather than three independent numerical update rates. Fig.~\ref{fig:update-cycle} summarizes this schedule.

At each eligible boundary, the implementation records rollout histories, credits the UCB arm associated with the evaluated policy, selects the next advice, updates best-so-far records, constructs \(\mathcal{B}^{k}\), and installs the sanitized LLM outputs before resetting the turtles. Network parameters, replay memory, \(\epsilon\), UCB state, policy histories, and lifetime statistics persist across rounds; only round-specific DQN action counts and loss statistics are cleared.

The round-level LLM output is a structured policy. Policy
$\boldsymbol{\theta}_i^k$ contains a numerical motion vector and four regional
action preferences,
\begin{equation}
\boldsymbol{\theta}_i^k
=\left(\mathbf{m}_i^k,\boldsymbol{\rho}_i^k\right),
\end{equation}
with
\begin{equation}
\begin{aligned}
\mathbf{m}_i^k
&=[\delta_i^k,\delta_{i,\mathrm{near}}^k,
\omega_i^k,\alpha_i^k,d_{i,\mathrm{near}}^k],\\
\boldsymbol{\rho}_i^k
&=[a_{i,\mathrm{FM}}^k,a_{i,\mathrm{FA}}^k,
a_{i,\mathrm{NM}}^k,a_{i,\mathrm{NA}}^k].
\end{aligned}
\end{equation}
Here, $\delta$ and $\delta_{\mathrm{near}}$ are the standard and near-goal
forward steps, $\omega$ is the maximum turn angle, $\alpha$ is the alignment
threshold, and $d_{\mathrm{near}}$ is the near-goal distance threshold. The
subscripts FM, FA, NM, and NA denote far/misaligned, far/aligned,
near/misaligned, and near/aligned regions.

Each regional preference belongs to the five-action set
\begin{equation}
\begin{aligned}
\mathcal{A}=\{&
\texttt{FORWARD},\texttt{FORWARD\_SMALL},\\
&\texttt{TURN\_ONLY},\texttt{TURN\_AND\_FORWARD},\\
&\texttt{TURN\_AND\_FORWARD\_SMALL}\}.
\end{aligned}
\end{equation}
For distance $d$ and signed goal-angle error $e$, the LLM-recommended action is
\begin{equation}
a_i^{\mathrm{LLM}}(d,e)=
\begin{cases}
a_{i,\mathrm{FM}}, & d>d_{i,\mathrm{near}},\
|e|>\alpha_i,\\
a_{i,\mathrm{FA}}, & d>d_{i,\mathrm{near}},\
|e|\leq\alpha_i,\\
a_{i,\mathrm{NM}}, & d\leq d_{i,\mathrm{near}},\
|e|>\alpha_i,\\
a_{i,\mathrm{NA}}, & d\leq d_{i,\mathrm{near}},\
|e|\leq\alpha_i.
\end{cases}
\end{equation}
This recommendation is a soft prior for the DQN, not a compulsory action.
The DQN-selected category is converted to motion using the active policy:
\texttt{FORWARD} and \texttt{FORWARD\_SMALL} translate by $\delta$ and
$\delta_{\mathrm{near}}$; \texttt{TURN\_ONLY} rotates toward the goal by
$\min(\omega,|e|)$; and the two turn-and-forward actions apply the same bounded
rotation before the corresponding translation.

The generated numerical fields are sanitized to the implementation ranges
\begin{equation}
\begin{aligned}
0.5&\leq\delta\leq3.0,&
0.01&\leq\delta_{\mathrm{near}}\leq1.0,\\
3&\leq\omega\leq60,&
1&\leq\alpha\leq45,\\
1&\leq d_{\mathrm{near}}\leq10.&&
\end{aligned}
\end{equation}
Here $\omega$ and $\alpha$ are measured in degrees. The parser
first attempts to recover either a JSON object or a Python dictionary from the
LLM response. Once a dictionary is recovered, policy sanitization starts from
\texttt{DEFAULT\_POLICY} and overwrites only the supplied fields that can be
validated; missing or invalid numerical and action fields therefore retain
their field-specific defaults. Failure to recover a dictionary invokes a
stage-dependent fallback. An initialization failure loads the complete default
policy, whereas a failure during a later round-level refinement retains the
robot's previously active control policy. This distinction prevents malformed
LLM output from replacing an established controller while guaranteeing an
executable policy before the first rollout.

At initialization, each LLM receives the environment description, its robot's
start state, the goal, the action vocabulary, and the output schema. For subsequent policy
refinement, $\mathcal{B}^k$ aggregates each robot's outcome and final state, current
policy fields, mission and rationale, expected DQN improvement, cross-LLM communication
summary, current and next UCB advice, and DQN round feedback. The subsequent
policy refinement is
\begin{equation}
\boldsymbol{\theta}_i^{k+1}
=\mathcal{S}\!\left[
F_{\mathcal{M}_i}
\!\left(
H_i^k,\boldsymbol{\theta}_i^k,
\mathcal{B}^k,z_i^{k+1}
\right)\right],
\end{equation}
where $F_{\mathcal{M}_i}$ denotes generation by robot $i$'s dedicated model,
$H_i^k$ is its complete local history, $z_i^{k+1}$ is its local UCB advice, and
$\mathcal{S}$ is the sanitization operator. Every model receives the same board
but only its own current policy, complete history, and next advice; the three
model calls run concurrently with three worker threads, and each returned policy
is loaded only by its corresponding robot. The output also records a mission
statement, rationale, expected DQN improvement, and cross-LLM learning summary
for the next board. These textual fields do not enter the tick-level action loop.

\subsection{UCB-Guided Policy Refinement}

UCB1 supplies a robot-local exploration--exploitation rule over five interpretable LLM refinement modes \cite{auer},
\begin{equation}
\begin{aligned}
\mathcal{Z}=\{&
\texttt{KEEP\_CURRENT},
\texttt{ADOPT\_BEST\_SHARED},\\
&\texttt{SMALL\_MUTATION},
\texttt{EXPLORE\_NEW},\\
&\texttt{HYBRID\_WITH\_BEST}\}.
\end{aligned}
\end{equation}
The modes ask the LLM, respectively, to retain the current policy, adopt useful components from the best shared policy, apply a small numerical mutation, explore a substantially different policy, or combine stable local components with useful shared components. An arm is natural-language refinement advice; it neither selects a motion action nor chooses an LLM backend. Untried arms are selected first in the fixed order listed above.

If arm \(z_i^k\) generated the policy evaluated in round \(k\), its count and empirical value are updated as
\begin{equation}
\begin{aligned}
n_{i,z_i^k}&\leftarrow n_{i,z_i^k}+1,\\
\overline{R}_{i,z_i^k}
&\leftarrow
\overline{R}_{i,z_i^k}
+\frac{R_i^k-\overline{R}_{i,z_i^k}}
{n_{i,z_i^k}}.
\end{aligned}
\end{equation}
Here, \(n_{i,z}\) is the number of credited rollouts and \(\overline{R}_{i,z}\) is the online empirical mean. The initial LLM policy is not associated with a UCB arm and is therefore not credited.

For a successful rollout, the implementation defines the optimistic movement lower bound
\begin{equation}
L_i=\max\!\left(1,
\left\lceil\frac{d_i^0}{\delta_{\max}}\right\rceil\right),
\qquad \delta_{\max}=3.0,
\end{equation}
where \(d_i^0\) is the fixed start-to-goal distance and \(\delta_{\max}\) is the largest sanitized forward step. This bound counts translation ticks without adding turns. The UCB credit is
\begin{equation}
R_i^k=\operatorname{round}_{6}\!\left(
\begin{cases}
0, & y_i^k=0,\\
\min\!\left(1,\dfrac{L_i}{\max(T_i^k,1)}\right), & y_i^k=1,
\end{cases}
\right),
\end{equation}
where \(\operatorname{round}_{6}\) is the code's six-decimal rounding operation. Thus, \(R_i^k\in[0,1]\): it is zero for failure, equals one at the optimistic lower bound, and decreases with successful completion time. Accumulated tick-level DQN reward is excluded because the bandit credits the round-level refinement mode rather than individual actions. The denominator clamp reproduces the code's protection against a zero finish-tick value.

After all arms have been observed, the next mode is selected by
\begin{equation}
z_i^{k+1}
=\arg\max_{z\in\mathcal{Z}}
\left[
\overline{R}_{i,z}+
c\sqrt{\frac{\ln(n_i+1)}{n_{i,z}}}
\right],
\qquad c=1.4,
\end{equation}
where \(n_i=\sum_{z\in\mathcal{Z}}n_{i,z}\). The evaluated implementation fixes \(c=1.4\) for every robot and does not tune it by LLM backend, configuration, or round. UCB therefore selects only the next refinement mode; numerical and categorical policy synthesis remains the responsibility of the robot's LLM.

\subsection{Policy-Conditioned Local Double DQN Control}

Each robot implements a Double DQN controller based on DQN and Double
Q-learning \cite{mnih,doubleq}. It owns an online network
$Q_i(\cdot;\boldsymbol{\phi}_i)$ and a target network
$Q_i^{-}(\cdot;\boldsymbol{\phi}_i^{-})$ with architecture
\begin{equation}
\mathbb{R}^{15}\rightarrow 64\ \mathrm{ReLU}
\rightarrow64\ \mathrm{ReLU}\rightarrow\mathbb{R}^{5}.
\end{equation}
The two hidden layers each contain 64 ReLU units, and the five outputs are the
Q-values for the five action categories in $\mathcal{A}$.
The 15-dimensional state concatenates navigation, active-policy, and LLM-prior
components:
\begin{equation}
s_{i,t}^k=
\left[
s_{i,t}^{\mathrm{nav}},
s_{i,t}^{\mathrm{policy}},
s_{i,t}^{\mathrm{prior}}
\right].
\end{equation}
Each block contributes five entries, yielding the 15-dimensional input used by
the implemented network.
With $d_t$ and $e_t$ denoting the current distance and signed angle error, the
navigation component is
\begin{equation}
s_{i,t}^{\mathrm{nav}}=
\left[
\frac{d_t}{100},
\frac{e_t}{180},
\mathbf{1}[d_t\leq d_{\mathrm{near}}],
\mathbf{1}[|e_t|\leq\alpha],
\frac{d_t}{100}
\right].
\end{equation}
The fifth entry reserves a distance-history channel within the fixed 15-dimensional interface. When the post-action next state is encoded, \texttt{rl-prev-distance} still stores the pre-action distance; the replayed transition therefore contains both the post-action distance in the first entry and its preceding value in the fifth entry, providing an explicit one-step progress signal for the Double-DQN update. The tracker is synchronized after the transition is stored, so the two entries coincide when the following online action is selected. Retaining this channel keeps the feature layout consistent across online action selection, replay storage, and minibatch training.
The policy component is
\begin{equation}
s_{i,t}^{\mathrm{policy}}=
\left[
\frac{\delta}{3.0},
\frac{\delta_{\mathrm{near}}}{1.0},
\frac{\omega}{60.0},
\frac{\alpha}{45.0},
\frac{d_{\mathrm{near}}}{10.0}
\right],
\end{equation}
and $s_{i,t}^{\mathrm{prior}}=\mathbf{p}_{i,t}\in\{0,1\}^{5}$ is the one-hot
encoding of $a_i^{\mathrm{LLM}}(d_t,e_t)$. Conditioning on the policy parameters
prevents the same categorical action from appearing identical to the DQN after
an LLM update changes its physical step or turn magnitude.

During training, the behavior policy is
\begin{equation}
a_{i,t}=
\begin{cases}
\operatorname{Uniform}(\mathcal{A}), & u<\epsilon_i,\\
\displaystyle
\arg\max_{a\in\mathcal{A}}
\left[
Q_i(s_{i,t},a;\boldsymbol{\phi}_i)
+\beta p_{i,t,a}
\right], & u\geq\epsilon_i,
\end{cases}
\end{equation}
where $u\sim\operatorname{Uniform}(0,1)$ and $\epsilon_i$ is the robot's current exploration probability. The evaluated C4 controller fixes $\beta=0.05$; the vanilla DQN controllers in C1 and C2 do not apply this additive prior bias. In C4, the bias guides exploitation without preventing learned Q-value differences from selecting another action.

For replay transition
$(s_{i,t},a_{i,t},r_{i,t},s_{i,t+1},\eta_{i,t})$, where $\eta_{i,t}=1$ for a
terminal transition, Double DQN uses the online network, including the same LLM
prior, to select the next action:
\begin{equation}
a_{i,t+1}^{*}
=\arg\max_{a\in\mathcal{A}}
\left[
Q_i(s_{i,t+1},a;\boldsymbol{\phi}_i)
+\beta p_{i,t+1,a}
\right].
\end{equation}
The target network evaluates that action,
\begin{equation}
\widehat{y}_{i,t}
=r_{i,t}
+\gamma(1-\eta_{i,t})
Q_i^{-}(s_{i,t+1},a_{i,t+1}^{*};
\boldsymbol{\phi}_i^{-}),
\qquad \gamma=0.99.
\end{equation}
The prior bias therefore affects online-network action selection in the
preceding equation, but it is not added to the target-network value gathered in
this equation.
For mini-batch $\mathcal{D}_B$, the online network minimizes Smooth L1 loss,
\begin{equation}
\mathcal{L}_i
=\frac{1}{|\mathcal{D}_B|}
\sum_{j\in\mathcal{D}_B}
\ell_{\mathrm{H}}
\!\left(
Q_i(s_j,a_j;\boldsymbol{\phi}_i)-\widehat{y}_j
\right),
\end{equation}
where
\begin{equation}
\ell_{\mathrm{H}}(x)=
\begin{cases}
\frac{1}{2}x^2, & |x|<1,\\
|x|-\frac{1}{2}, & |x|\geq1.
\end{cases}
\end{equation}
Thus, $x=Q_i(s_j,a_j;\boldsymbol{\phi}_i)-\widehat{y}_j$ is the temporal-
difference residual, and the implementation averages this Smooth L1 loss over
the sampled mini-batch.

For transition \((s_t,a_t,r_t,s_{t+1})\), let \((d_t,e_t)\) and \((d_{t+1},e_{t+1})\) denote the pre-action and post-action distance and signed angle error. The implemented reward is the sum of independently activated components,
\begin{equation}
r_t=-0.01+r_t^{\mathrm{away}}
+r_t^{\mathrm{near\text{-}away}}
+r_t^{\mathrm{turn}}
+r_t^{\mathrm{large}}
+r_t^{\mathrm{success}}
+r_t^{\mathrm{timeout}}.
\end{equation}
Table~\ref{tab:tick-reward} gives every component used by the code, with \(\varepsilon_d=\varepsilon_e=10^{-6}\). Here, \(\sigma_{t+1}\) is the post-action exact-goal indicator and \(\varphi_{t+1}=\mathbf{1}[t+1\geq T_{\max}]\) is the timeout flag passed to the DQN.

\begin{table}[t]
\caption{Implemented Tick-Level Reward Components}
\label{tab:tick-reward}
\centering
\scriptsize
\setlength{\tabcolsep}{3pt}
\resizebox{\columnwidth}{!}{%
\begin{tabular}{llc}
\hline
Component & Activation condition & Value \\
\hline
Base & Every executed transition & \(-0.01\) \\
Away & \(d_{t+1}>d_t+\varepsilon_d\) & \(-0.02\) \\
Near-away & \(d_t\leq d_{\mathrm{near}}\) and away & \(-0.05\) \\
Corrective turn & \(a_t=\texttt{TURN\_ONLY}\), \(|e_t|>\alpha\), & \(+0.005\) \\
 & \(|e_{t+1}|<|e_t|-\varepsilon_e\) & \\
Other turn-only & \(\texttt{TURN\_ONLY}\) and corrective condition false & \(-0.01\) \\
Near large motion & \(d_t\leq d_{\mathrm{near}}\), \(\sigma_{t+1}=0\), & \(-0.01\) \\
 & \(a_t\in\{\texttt{FORWARD},\texttt{TURN\_AND\_FORWARD}\}\) & \\
Success & \(\sigma_{t+1}=1\) & \(+1.0\) \\
Timeout & \(\varphi_{t+1}=1\) & \(-1.0\) \\
\hline
\end{tabular}%
}
\end{table}

Because the away and near-away tests are independent, both penalties apply when a robot moves away from the goal inside the near-goal region. A nonterminal corrective turn with no other activated penalty has net reward \(-0.005\) after the base cost. Success and timeout are also added independently. If exact-goal arrival and timeout occur on the same transition, their terminal contributions cancel, while the base and any other activated components remain. 

Training uses Adam with learning rate $10^{-3}$, batch size 32, replay capacity
5000, and target-network synchronization every 100 successful training
updates. After each mini-batch update,
\begin{equation}
\epsilon_i\leftarrow
\max(0.05,0.997\,\epsilon_i),
\qquad \epsilon_i^{0}=1.0.
\end{equation}
Thus, exploration starts at one, decays by a factor of $0.997$ after each
successful mini-batch update, and is lower-bounded by $0.05$. Learning begins
when the replay buffer contains at least 32 transitions.

Algorithm~\ref{alg:framework} gives the end-to-end procedure for the evaluated
configuration.
\begin{algorithm}[t]
\caption{Two-Level LLM--UCB--DQN Procedure}
\label{alg:framework}
\footnotesize
\begin{algorithmic}[1]
\STATE Initialize one policy with each dedicated LLM in parallel.
\STATE Initialize independent UCB1 and Double DQN states for all robots.
\FOR{$k=1,\ldots,K$}
    \STATE Reset turtle poses and round-specific DQN statistics.
    \WHILE{at least one robot is active and $t<T_{\max}$}
        \FOR{each active robot $r_i$}
            \STATE Encode $s_{i,t}^{k}$ and select $a_{i,t}^{k}$.
            \STATE Execute $a_{i,t}^{k}$ using $\boldsymbol{\theta}_i^k$.
            \STATE Compute $r_{i,t}^{k}$ and observe $s_{i,t+1}^{k}$.
            \STATE Store the transition and train local Double DQN.
        \ENDFOR
        \STATE Advance the NetLogo tick.
    \ENDWHILE
    \STATE Record $y_i^k$, $T_i^k$, local history, and DQN feedback.
    \STATE Update the previously selected UCB arm, if one exists.
    \STATE Select $z_i^{k+1}$ for every robot; update each successful robot's
    \texttt{best-tick} and \texttt{best-policy} records.
    \IF{$k<K$}
        \STATE Construct $\mathcal{B}^k$ from all robot summaries.
        \STATE Generate and sanitize all
        $\boldsymbol{\theta}_i^{k+1}$ in parallel.
    \ENDIF
\ENDFOR
\STATE Return finish-tick trajectories and persistent learning statistics.
\end{algorithmic}
\end{algorithm}

\section{Experimental Design}

\subsection{Evaluation Objectives}
The evaluation addresses four questions. \textbf{RQ1} examines the joint effect of cross-LLM communication and UCB refinement-mode selection. \textbf{RQ2} compares learned tick-level local control with direct execution of LLM-generated regional rules under the same round-level components. \textbf{RQ3} compares the policy-conditioned Double DQN in C4 with the state-only vanilla DQN in C2. \textbf{RQ4} characterizes round-level adaptation and robot-specific outcomes in C1 and C4. Because each LLM backend remains assigned to the same robot, robot-wise results describe the instantiated composite heterogeneous robots and do not rank the backends.

\subsection{Controller Configurations}
Table~\ref{tab:experimental-configurations} defines the four configurations. Each retains one dedicated LLM per robot and follows the round-level schedule in Section~\ref{sec:round_update}. The world, start and goal assignments, action vocabulary, and motion interface are fixed; the controller and learning differences are stated below.

\begin{table}[t]
\caption{Controller Configurations}
\label{tab:experimental-configurations}
\centering
\footnotesize
\setlength{\tabcolsep}{2.2pt}
\resizebox{\columnwidth}{!}{%
\begin{tabular}{lccccc}
\hline
Configuration & Shared board & UCB & Tick-level control & Input dim. & Prior bias \\
\hline
C1: Local LLM--DQN & No & No & Vanilla DQN & 5 & No \\
C2: Cross-LLM--UCB + DQN & Yes & Yes & Vanilla DQN & 5 & No \\
C3: Cross-LLM--UCB, no DQN & Yes & Yes & Regional rule & -- & -- \\
C4: Complete configuration & Yes & Yes & Double DQN & 15 & 0.05 \\
\hline
\end{tabular}%
}
\end{table}

C1 uses robot-local history for subsequent policy refinement and disables the shared board and UCB; its vanilla DQN receives the five-dimensional navigation observation. C2 enables cross-LLM communication and UCB refinement-mode selection while retaining the same vanilla DQN. C3 keeps the round-level LLM--UCB components and executes the current policy's regional action preference directly, without DQN. The complete configuration C4 uses the 15-dimensional policy-conditioned Double DQN described in Section~III.

C1--C2 changes cross-LLM communication and UCB jointly, so it cannot separate their effects. C2--C3 removes the learned tick-level controller. C1 and C2 use a five-dimensional vanilla DQN with $\gamma=0.95$, MSE loss, an $\epsilon$-decay factor of $0.9995$, and the distance-progress reward implemented in those configurations. C4 uses the 15-dimensional Double DQN with $\gamma=0.99$, Smooth L1 loss, an $\epsilon$-decay factor of $0.997$, the reward in Table~\ref{tab:tick-reward}, and the additive LLM-prior bias $\beta=0.05$. C2--C4 is therefore a bundled configuration comparison rather than an isolated test of Double DQN.

\subsection{Common Simulation Setup}
The NetLogo world spans $[-50,50]\times[-50,50]$, and the common goal is located at $(10,-5)$. The three robots start at $(49,40)$, $(-29,40)$, and $(-29,-50)$ and initially face the goal. They are respectively paired with Llama-3.3-70B-Instruct-quantized, Phi-4, and Meta-Llama-3.1-70B-Instruct-quantized. These assignments, the five-action motion interface, and all physical transition rules are fixed across configurations.

Each configuration is evaluated for 30 rounds, with at most 1000 ticks per round. This produces 90 robot--round records (three robots over 30 rounds). Because policy and learning states persist across rounds, these records are analyzed as repeated measurements rather than independent replicates. The round-level LLM calling schedule follows Section~\ref{sec:round_update}. A robot's round ends when it reaches the goal patch or exhausts the tick budget.

\subsection{LLM Backends and Simulation View}

The three dedicated policy agents use Llama-3.3-70B-Instruct, Phi-4, and Llama-3.1-70B-Instruct. Llama 3.1 and 3.3 belong to Meta's instruction-tuned Llama 3 family, which supports multilingual text, coding, reasoning, and tool-use tasks~\cite{llama3herd}. The Llama-3.3 model card specifies a 70B text model with a 128k-token context window, optimized for multilingual dialogue~\cite{llama33card}. Phi-4 is a 14B model whose training emphasizes high-quality data, synthetic reasoning data, and post-training; its technical report also identifies strict instruction following as a limitation~\cite{phi4report}. These published characteristics motivate using distinct backends for robot-specific policy generation, but they do not predict navigation performance directly. The evaluated endpoints include service-specific quantized Llama variants, and the exact serving stack and quantization parameters are not exposed by the current interface.

\begin{figure}[!t]
\centering
\includegraphics[width=\columnwidth]{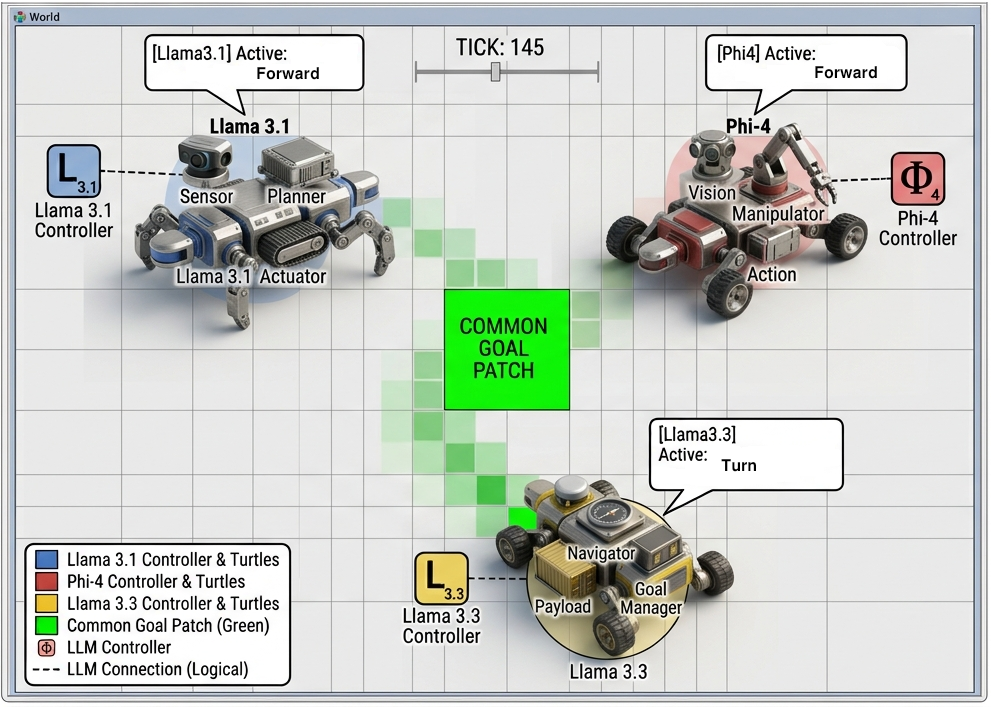}
\caption{Representative NetLogo rollout with three composite heterogeneous robots and the common goal patch. Text labels show the active runtime action and local best-tick record for that displayed state.Each robot is connected to an LLM, specifically, LLama3.1, Phi4, Llama3.3 respectively.}
\label{fig:netlogo_snapshot}
\end{figure}

Figure~\ref{fig:netlogo_snapshot} shows a representative NetLogo rollout. The colored turtles are paired with their dedicated LLM agents, and the green patch is the common goal. On-screen action and best-tick labels are runtime status indicators, not aggregate experimental statistics. Because each model is permanently paired with one start pose and one local learner, later agent-level differences are interpreted as properties of the complete instantiated robot--controller combinations rather than strengths or weaknesses of the isolated LLMs.

\subsection{Evaluation Protocol and Metrics}
At the start of every round, robot positions, headings, and round-local counters are reset to the common initial conditions. The enabled UCB statistics, DQN learner state, policy history, and best-policy record persist across rounds. The policy generated at the round boundary remains fixed during the ensuing rollout, while the tick controller defined in Table~\ref{tab:experimental-configurations} produces the physical actions.

For each robot--round observation, the simulator records goal attainment and the successful completion tick. The primary metrics are success rate and the distribution of successful completion ticks, summarized by the median, interquartile range (IQR), mean, and 90th percentile (P90). Round-wise traces and early-to-late medians describe adaptation over the fixed horizon. Two complementary descriptive metrics combine success with tick efficiency and retain slow-tail behavior. Episode reward, exploration rate, action counts, and training loss are retained only as DQN diagnostics. LLM computation cost is outside the scope of this draft.

\subsection{Primary Metrics and Analysis}
\label{subsec:primary_metrics}

Let \(y_{i,c}^{k}\in\{0,1\}\) indicate whether robot \(i\) in configuration \(c\) reaches the goal in round \(k\). With \(N=3\) robots and \(K=30\) rounds, the task-success rate is
\begin{equation}
\mathrm{SR}_{c}=\frac{1}{NK}\sum_{i=1}^{N}\sum_{k=1}^{K}y_{i,c}^{k}.
\label{eq:success_rate}
\end{equation}
The value \(T_{\max}=1000\) is a timeout code rather than a successful completion tick. The data contain one such event, for the Llama-3.1 composite robot under C2 in round 12. Success rate is reported over all 90 robot--round records for each configuration, while completion-time summaries use only successful records. Because within-configuration records share persistent policy and learning states, these summaries are descriptive rather than estimates based on independent trials.

Because the successful finish-tick distributions are right-skewed, the primary efficiency statistic is the median with the interquartile range (IQR). To describe adaptation without selecting a favorable round, the round-wise trace uses the median of the three recorded values, with a timeout retained at \(T_{\max}\). Early and late performance pool the successful observations from rounds 1--5 and 26--30, respectively. Their relative change is
\begin{equation}
I_c=100\,
\frac{M^{\mathrm{early}}_c-M^{\mathrm{late}}_c}
     {M^{\mathrm{early}}_c},
\label{eq:early_late_improvement}
\end{equation}
where a positive \(I_c\) denotes fewer ticks in the final five rounds. Because policy and learning states persist across rounds, this quantity describes change over the evaluation horizon and is not an inferential effect estimate.

Following the success-weighted time-step metric used in navigation evaluation~\cite{wang2024interactive}, lower-bound-normalized success efficiency (LBSE) is
\begin{equation}
\mathrm{LBSE}_c=\frac{1}{NK}\sum_{i=1}^{N}\sum_{k=1}^{K}
y_{i,c}^{k}\min\!\left(1,\frac{L_i}{\max(T_{i,c}^{k},1)}\right),
\label{eq:lbse}
\end{equation}
where \(L_i=\lceil d_i^0/3\rceil\) is the optimistic movement lower bound already used by the round-level UCB reward. The fixed start coordinates give \(L_i=20\) for all three robots. Hence, \(\mathrm{LBSE}_c\in[0,1]\), with larger values indicating more successful completions closer to this lower bound. Because it averages the same efficiency signal used internally by UCB, LBSE is an implementation-aligned diagnostic rather than independent validation of the bandit objective.

To retain rare slow outcomes, we also report the empirical worst-decile completion cost, a finite-sample upper-tail summary motivated by risk-sensitive RL practice~\cite{ni2024risk}. Let \(T_{c,(1)}\leq\cdots\leq T_{c,(NK)}\) be all tick costs in configuration \(c\), including a timeout as \(T_{\max}\), and let \(m=NK-\lfloor0.90NK\rfloor=9\). Then
\begin{equation}
\mathrm{WDC}_{0.10,c}=\frac{1}{m}
\sum_{j=NK-m+1}^{NK}T_{c,(j)}.
\label{eq:worst_decile_cost}
\end{equation}
Lower values indicate a smaller average cost among the slowest 10\% of the 90 observations. This statistic is a finite-sample upper-tail mean, not an estimate of population CVaR. LBSE and \(\mathrm{WDC}_{0.10,c}\) remain descriptive summaries of the recorded configuration-level data; neither supplies an independent-sample significance test.

\section{Results and Analysis}
\label{sec:results}

\begin{figure*}[!t]
\centering
\begin{minipage}[t]{0.48\textwidth}
  \centering
  \includegraphics[width=\linewidth]{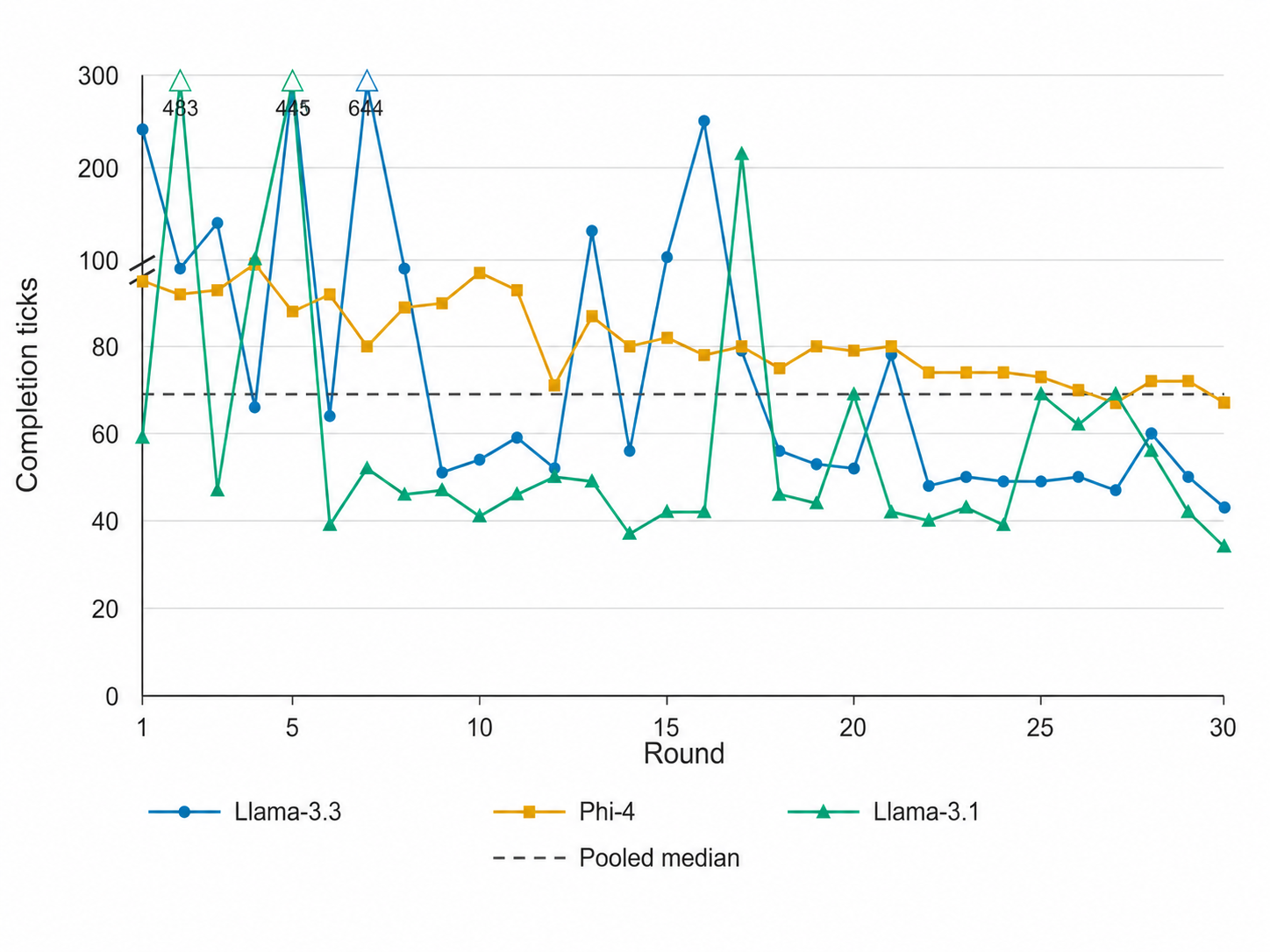}\\[-0.5ex]
  \footnotesize\textbf{(a)} C1: LLM (no shared board) + DQN.
\end{minipage}
\hfill
\begin{minipage}[t]{0.48\textwidth}
  \centering
  \includegraphics[width=\linewidth]{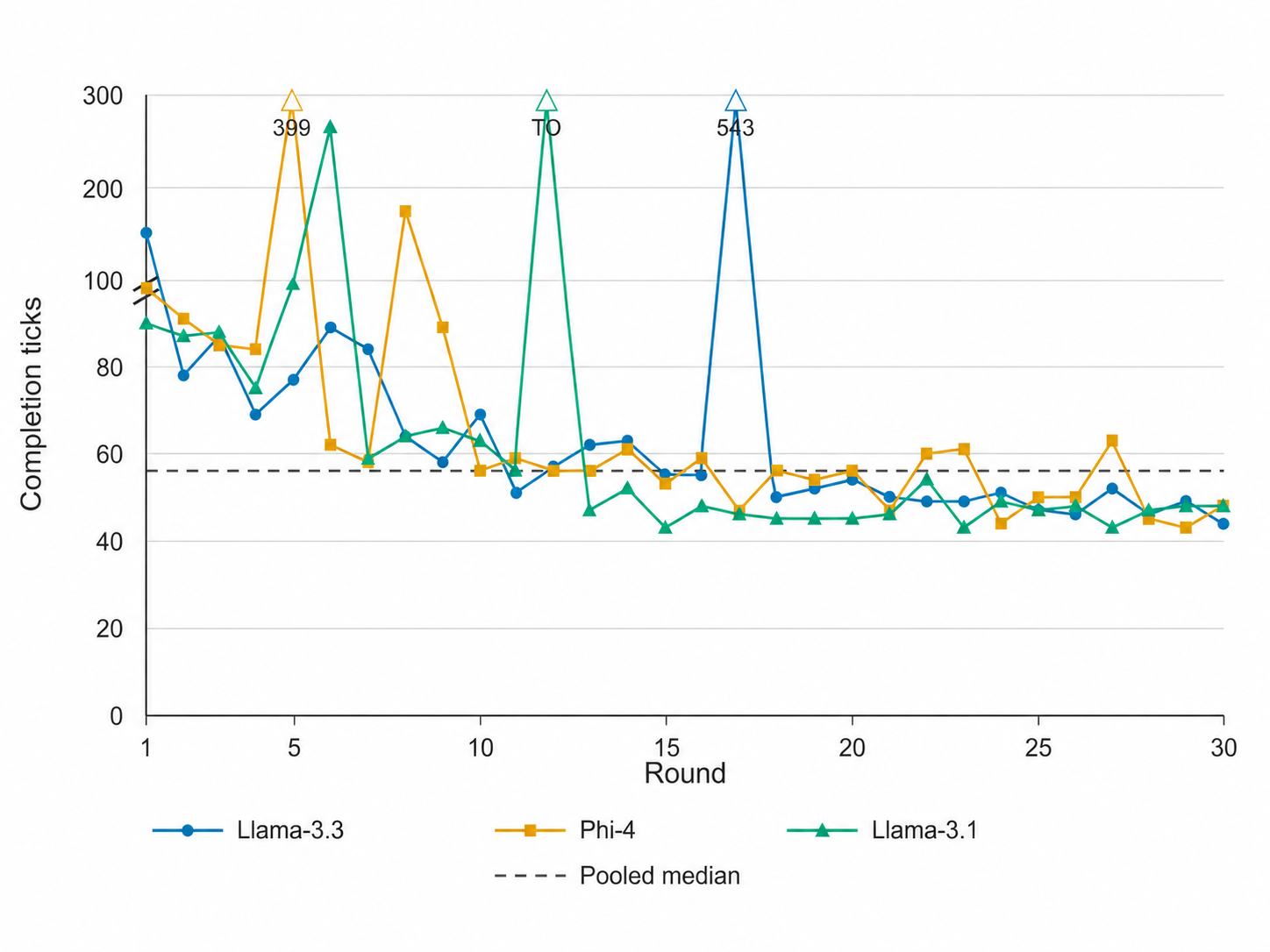}\\[-0.5ex]
  \footnotesize\textbf{(b)} C2: LLM with shared board communication + UCB + DQN.
\end{minipage}

\vspace{0.7ex}
\begin{minipage}[t]{0.48\textwidth}
  \centering
  \includegraphics[width=\linewidth]{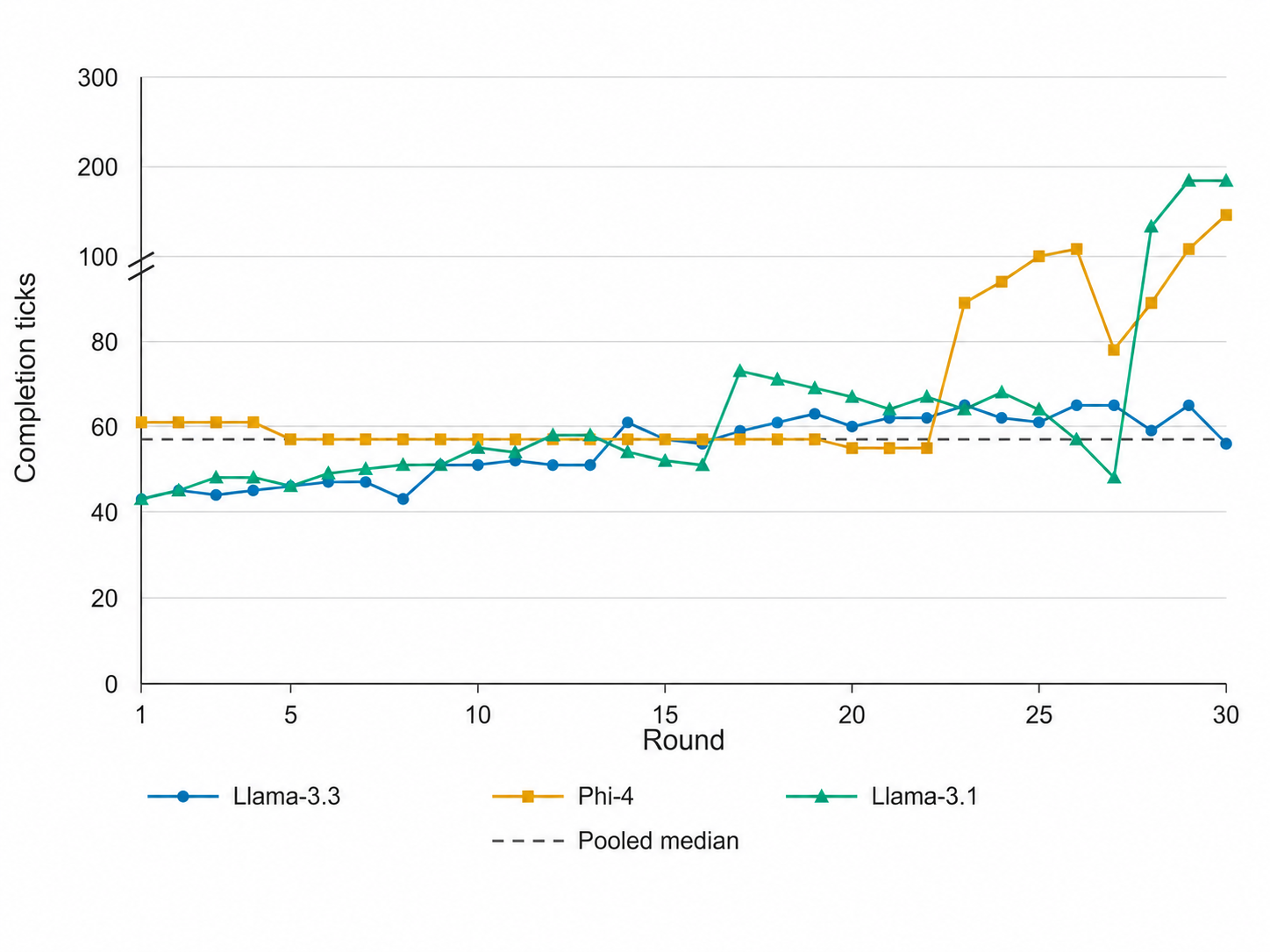}\\[-0.5ex]
  \footnotesize\textbf{(c)} C3: LLM with shared board communication + UCB but without DQN.
\end{minipage}
\hfill
\begin{minipage}[t]{0.48\textwidth}
  \centering
  \includegraphics[width=\linewidth]{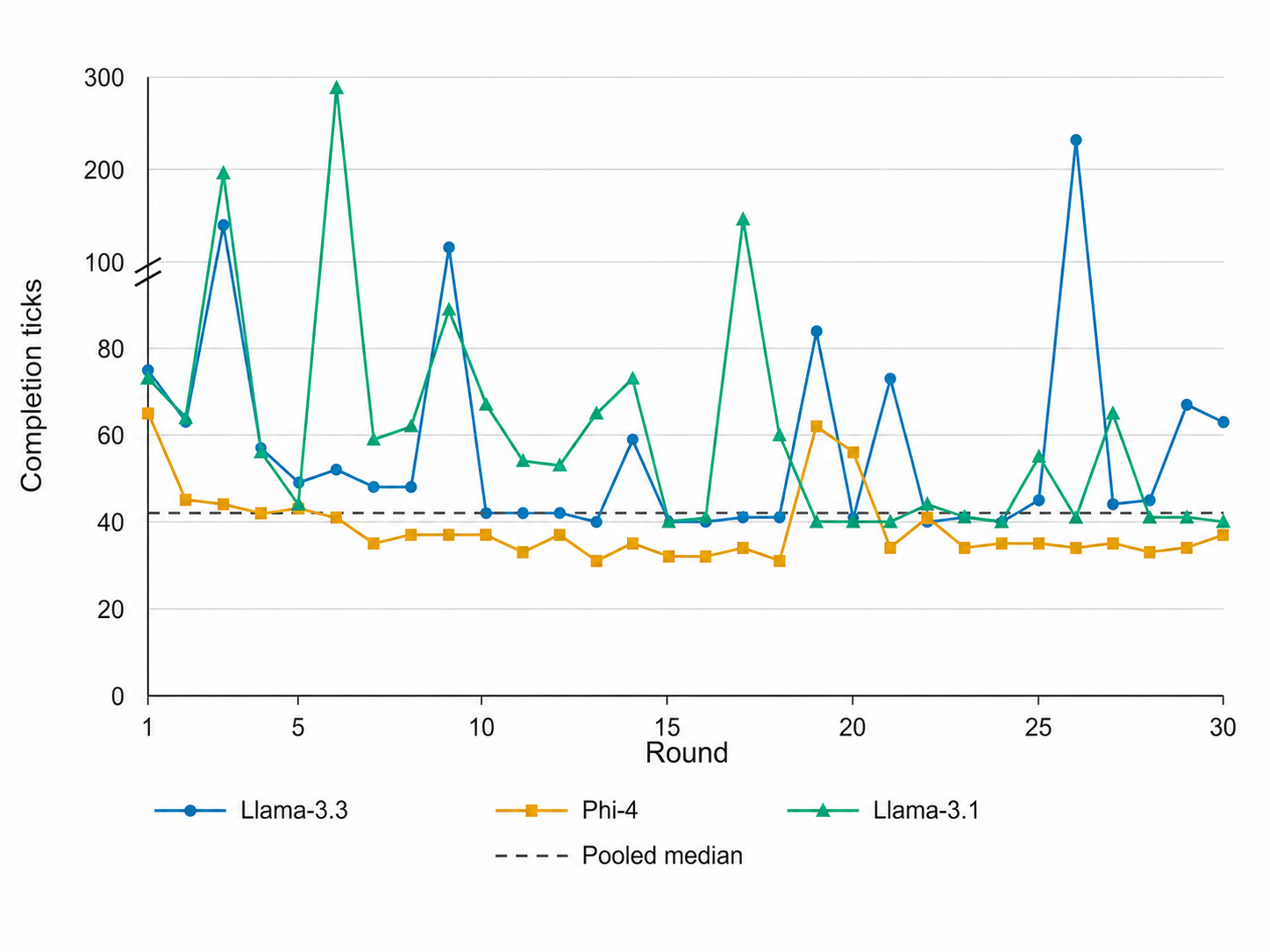}\\[-0.5ex]
  \footnotesize\textbf{(d)} C4: LLM with shared board communication + UCB + Double DQN.
\end{minipage}
\caption{Completion ticks over 30 rounds for the four bundled configurations. Each curve corresponds to one fixed composite heterogeneous robot, and the dashed line is the pooled median. All panels use the same piecewise-linear vertical scale, which expands 0--100 ticks and compresses 100--300 ticks. The break marks the change of scale; values above 300 ticks are placed at the upper boundary and labeled with their actual values, while the C2 timeout is marked as \texttt{TO}.}
\label{fig:config_trajectories}
\end{figure*}

The four configurations are reported as bundled system ablations. To expose the progression requested for the analysis, this section first examines the local LLM--DQN baseline (C1), then the two cross-LLM configurations with vanilla DQN (C2) or no DQN (C3), and finally the complete policy-conditioned Double DQN configuration (C4). The configurations do not form a strict single-factor sequence: C1 already uses round-level robot-local LLM refinement, and C2--C4 differ in more than one implementation choice. The results therefore support configuration-level observations rather than isolated causal effects. Figure~\ref{fig:config_trajectories} uses the same disclosed piecewise-linear vertical scale: the 0--100-tick interval occupies 70\% of the plot height, while 100--300 ticks occupy the remaining 30\%. The axis break at 100 ticks marks the change of scale, and observations above 300 ticks retain their actual numerical labels.

\subsection{C1: Local LLM--DQN Baseline}

C1 disables the shared board and UCB while retaining one robot-local LLM and a five-dimensional vanilla DQN. All 90 robot--round records were successful. Its successful completion-tick median was 69, with a mean of 91.5 and a P90 of 105.8 ticks. Figure~\ref{fig:config_trajectories}(a) shows that the early rounds contain several high-tick observations, including the largest successful value in the dataset (644 ticks). The pooled median decreased from 98 ticks in rounds 1--5 to 60 ticks in rounds 26--30. Thus, the local history and DQN state were sufficient to produce a faster late segment, but the configuration retained a pronounced upper tail.

\subsection{C2: Cross-LLM Communication, UCB, and Vanilla DQN}

C2 enables the shared board and robot-local UCB refinement-mode selection while retaining the five-dimensional vanilla DQN. It recorded 89 successes and one timeout, corresponding to a success rate of 98.9\%. Among successful observations, the median was 56 ticks, the mean was 72.7, and the P90 was 89.2. The early and late medians were 87 and 48 ticks, respectively. As shown in Fig.~\ref{fig:config_trajectories}(b), most observations moved toward the lower-tick region, although isolated slow rounds remained and the Llama-3.1 composite robot timed out in round 12. Relative to C1, the lower median is associated with the combined addition of cross-LLM communication and UCB; the present configuration does not separate those two components.

\subsection{C3: Cross-LLM Communication and UCB Without DQN}

C3 preserves the round-level cross-LLM and UCB components but executes the current regional rule directly. All 90 robot--round records for C3 reached the goal. Its pooled median was 57 ticks, close to C2, while its mean and P90 were 64.0 and 89.0 ticks. Its temporal profile differed from the DQN-enabled configurations: the early median was 46 ticks, but the late median increased to 78 ticks. Figure~\ref{fig:config_trajectories}(c) shows that the slowest observations are concentrated near the end of the horizon, with final-round values of 56, 146, and 184 ticks for the three robots. The no-DQN configuration was therefore competitive in the pooled median but did not exhibit the same late-round reduction observed in C1, C2, and C4.

\begin{figure}[!t]
\centering
\includegraphics[width=\columnwidth]{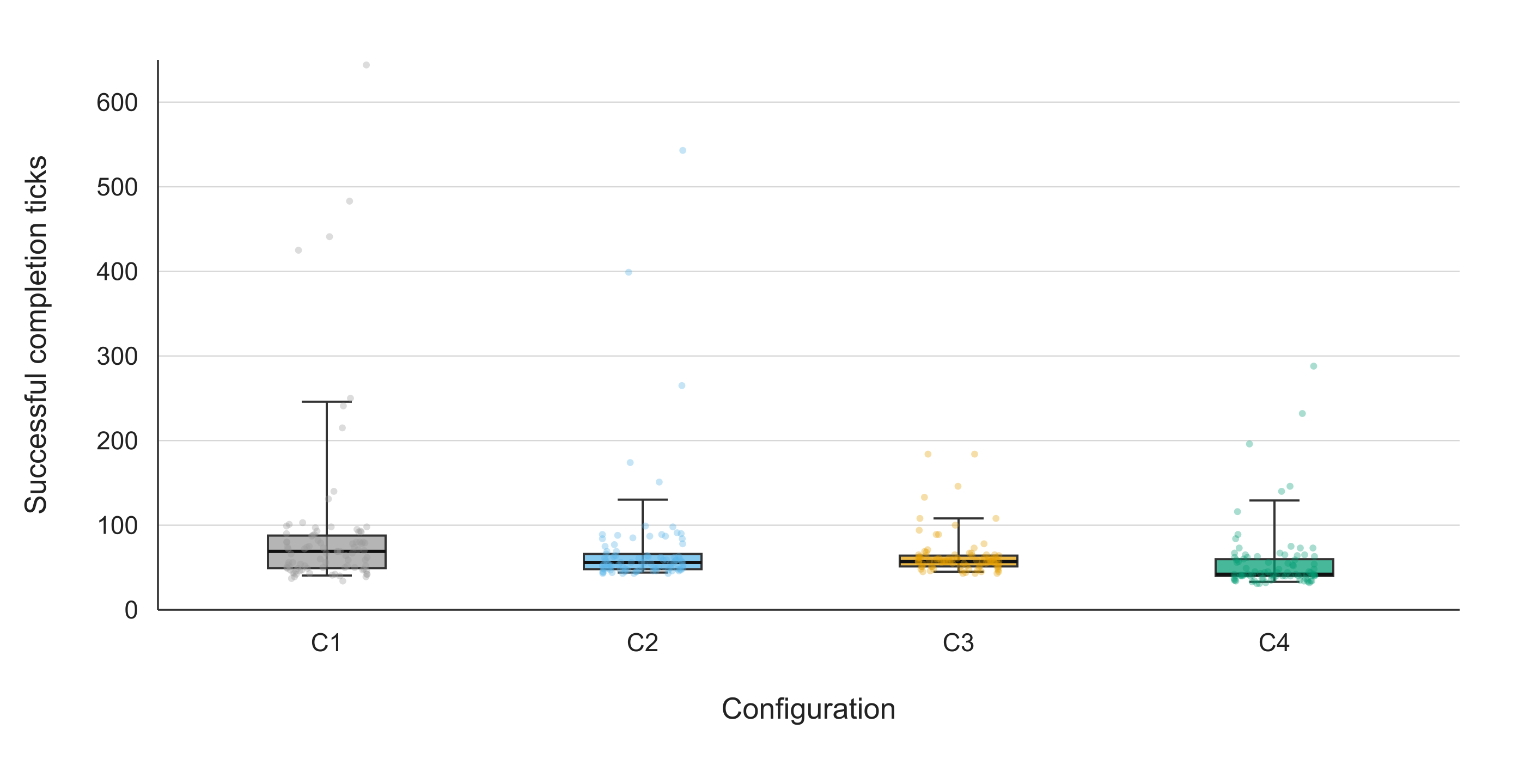}
\caption{Successful robot--round completion ticks by configuration. Boxes span the IQR, horizontal bars show the median, whiskers span the 5th--95th percentiles, and lightly jittered points show individual successful observations. The C2 timeout is excluded here and retained in Table~\ref{tab:overall_results} and the worst-decile cost.}
\label{fig:configuration_comparison}
\end{figure}

\begin{table}[!t]
\centering
\caption{Overall performance over 30 rounds. Completion-time statistics exclude timeouts; P90 is the 90th percentile of successful completion ticks. Early and late medians pool successful observations from rounds 1--5 and 26--30.}
\label{tab:overall_results}
\scriptsize
\setlength{\tabcolsep}{2.2pt}
\renewcommand{\arraystretch}{1.05}
\begin{tabular}{@{}lcccc@{}}
\hline
Metric & C1 & C2 & C3 & C4\\
\hline
Success (SR \%) & \shortstack{90/90\\(100.0)} & \shortstack{89/90\\(98.9)} & \shortstack{90/90\\(100.0)} & \shortstack{\textbf{90/90}\\\textbf{(100.0)}}\\
Median [IQR] & \shortstack{69.0\\\mbox{[49.3,87.8]}} & \shortstack{56.0\\\mbox{[48.0,66.0]}} & \shortstack{57.0\\\mbox{[51.3,64.0]}} & \shortstack{\textbf{42.0}\\\textbf{[40.0,59.8]}}\\
Mean & 91.5 & 72.7 & 64.0 & \textbf{56.3}\\
P90 & 105.8 & 89.2 & 89.0 & \textbf{73.2}\\
Early & 98 & 87 & 46 & 57\\
Late & 60 & 48 & 78 & \textbf{41}\\
\(I_c\) (\%) & 38.8 & 44.8 & \(-69.6\) & 28.1\\
\hline
\end{tabular}
\end{table}

\subsection{C4: Complete Multi-Timescale Configuration}

C4 combines cross-LLM communication, robot-local UCB advice, and policy-conditioned local Double DQN control. All 90 robot--round records for C4 reached the goal, and the configuration had the lowest observed median (42 ticks), mean (56.3 ticks), and P90 (73.2 ticks). Its early median of 57 ticks decreased to 41 ticks in the final five rounds. Figure~\ref{fig:config_trajectories}(d) also exposes non-monotonic behavior: the Llama-3.3 and Llama-3.1 robots contain isolated values of 232 and 288 ticks, respectively. The Phi-4 composite robot had the lowest within-configuration median (35 ticks) and the smallest IQR (7 ticks). These are agent-level observations under fixed start poses, policies, and learners; they do not establish an intrinsic ranking of the three LLM backends.

\subsection{Cross-Configuration Comparison}

Table~\ref{tab:overall_results} and Fig.~\ref{fig:configuration_comparison} compare the four bundled ablations over the complete horizon. C4 reduced the successful-observation median by 39.1\%, 25.0\%, and 26.3\% relative to C1, C2, and C3, respectively. Its P90 was 17.8\% below the next-lowest value (89.0 ticks in C3), indicating that its difference was present in the upper portion of the observed distribution as well as at the median. The success rates differ by only 1.1 percentage points: C1, C3, and C4 reached 90/90 goals, whereas C2 reached 89/90. Because the success-rate bars would add little information beyond Table~\ref{tab:overall_results}, Fig.~\ref{fig:configuration_comparison} displays only the completion-tick distributions. C2 and C3 had similar medians, but their early-to-late trajectories moved in opposite directions. This distinction would be hidden by a best-round or pooled-median-only comparison.

\begin{table}[tbp]
\centering
\caption{Complementary efficiency and slow-tail summaries.}
\label{tab:complementary_metrics}
\setlength{\tabcolsep}{6pt}
\begin{tabular}{lcc}
\hline
Config. & LBSE $\uparrow$ & $\mathrm{WDC}_{0.10}$ (ticks) $\downarrow$\\
\hline
C1 & 0.310 & 330.0\\
C2 & 0.339 & 313.3\\
C3 & 0.339 & \textbf{127.3}\\
C4 & \textbf{0.431} & 151.8\\
\hline
\end{tabular}
\end{table}

Table~\ref{tab:complementary_metrics} separates typical success-weighted efficiency from rare slow behavior. C4 had the highest LBSE, 39.0\%, 27.3\%, and 27.2\% above C1, C2, and C3, respectively. C3 instead had the lowest worst-decile cost. Its value of 127.3 ticks was 16.1\% below C4's 151.8 ticks because C4 retained isolated 232- and 288-tick rounds. The complete configuration therefore had the strongest aggregate normalized efficiency in this run, but it did not dominate every slow-tail criterion.

\begin{table}[tbp]
\centering
\caption{Successful completion-tick medians by composite robot.}
\label{tab:robot_results}
\setlength{\tabcolsep}{3.5pt}
\begin{tabular}{lccc}
\hline
Config. & Llama-3.3 & Phi-4 & Llama-3.1\\
\hline
C1 & 57.5 & 80.0 & \textbf{46.5}\\
C2 & 55.0 & 57.0 & 48.0\(^{\dagger}\)\\
C3 & 56.5 & 57.0 & 56.0\\
C4 & \textbf{46.5} & \textbf{35.0} & 54.5\\
\hline
\end{tabular}
\vspace{2pt}

\parbox{\columnwidth}{\footnotesize \(^{\dagger}\)Median over 29 successful observations; the remaining observation timed out.}
\end{table}

The agent-level medians in Table~\ref{tab:robot_results} show that the aggregate C4 ordering was not uniform: C1 had the lowest median for the Llama-3.1 composite robot. Each column combines a fixed LLM backend, start pose, policy history, and local controller state. Accordingly, the table characterizes the instantiated composite heterogeneous robots rather than the isolated language models.

The comparisons answer the evaluation questions only at the configuration level. C1--C2 changes communication and UCB jointly. C2--C3 removes learned tick-level control, but C3 also executes a different action-selection mechanism. C2--C4 changes the state representation, Q-learning rule, discount factor, loss, exploration schedule, reward, and prior bias. The four groups are therefore informative ablations of implemented systems, not single-factor estimates.

\subsection{Evidence Scope}

The reported results are descriptive configuration-level summaries. The 90 robot--round records for each configuration share persistent policies, UCB statistics, replay buffers, network weights, exploration states, and shared-board information across rounds, and are therefore treated as repeated measurements rather than independent replicates. These results do not support significance tests or causal attribution. Their scope is limited to the fixed NetLogo world, start assignments, goal, and LLM-to-robot mapping in Section~IV-C.
\FloatBarrier

\section{Conclusion}

This paper presented a decentralized multi-timescale system for composite heterogeneous robots, combining robot-local policy ownership, UCB-guided round-level refinement, and tick-level control. Cross-LLM communication is performed at round boundaries to support robot-specific policy refinement, whereas physical actions are selected at the tick level without repeated LLM calls. In the complete configuration, the local Double DQN conditions its value estimates on navigation state, the active LLM-generated policy, and the corresponding regional-action prior. Reinforcement learning therefore assists the LLM policy layer while leaving policy generation and robot-specific reasoning at the round level.

Across the four configurations, C4 combined complete observed success with the lowest pooled median and P90, indicating better aggregate central and upper-tail completion behavior in the fixed task. C3 remained competitive in pooled median completion time but worsened over the evaluation horizon, whereas agent-level medians were not uniformly lowest under C4. The strongest observed advantage was therefore configuration-level rather than uniform across robots or rounds.

The 90 robot--round records for each configuration share persistent policy and learning states across rounds and are therefore interpreted as repeated measurements rather than independent replicates. The four configurations also contain bundled implementation differences, so the present results neither isolate the causal contribution of every module nor demonstrate generalization to other environments or physical robots.

\subsection{Future Work}

Future experiments will introduce independent seeded repetitions and separate evaluation rollouts to quantify uncertainty independently of ongoing exploration. Single-factor ablations will isolate cross-LLM communication, UCB advice, the policy-conditioned state, Double DQN, reward design, and the LLM-prior bias. An explicit RL-only baseline and a direct per-tick LLM-control baseline will complete the progression from local reinforcement learning to round-level LLM assistance and cross-LLM communication. The fixed UCB coefficient will be subjected to sensitivity analysis.

The evaluation will also record LLM latency, token consumption, response validity, and fallback frequency. Further studies will vary LLM-to-start assignments, communication quality, obstacles, and task geometry, and will test physically heterogeneous platforms. These extensions are required before making claims about statistical reliability, computational efficiency, scalability, or real-world deployment.

\end{document}